\documentclass[letterpaper, 10 pt, conference]{ieeeconf}  
\IEEEoverridecommandlockouts                              
\usepackage{wrapfig,amsmath,amsthm,thmtools,amssymb,bm}
\usepackage{algorithm,tabularx}
\usepackage{algpseudocode}
\usepackage{graphicx}
\usepackage{epsfig} 
\usepackage{times} 
\usepackage{mathptmx} 
\usepackage{caption}
\usepackage{subcaption}

\newtheorem*{remark}{Remark}

\makeatletter
\newcommand{\multiline}[1]{%
  \begin{tabularx}{\dimexpr\linewidth-\ALG@thistlm}[t]{@{}X@{}}
    #1
  \end{tabularx}
}
\makeatother

\title{\LARGE \bf
Integrated Path Planning and Tracking Control of Marine Current Turbine in Uncertain Ocean Environments}

\author{Arezoo~Hasankhani,~\IEEEmembership{Student Member,~IEEE,}
        Ertugrul Baris Ondes,
        Yufei~Tang,~\IEEEmembership{Member,~IEEE,}\\
        Cornel Sultan, and
        James~VanZwieten
\thanks{This work was supported in part by the U.S. National Science Foundation under Grant Nos. ECCS-1809164, ECCS-1809404, \& OAC-2017597 and the U.S. Department of Energy under Grant No. DE-EE0008955.}
\thanks{A. Hasankhani and Y. Tang are with the Department of Computer \& Electrical Engineering and Computer Science, Florida Atlantic University, Boca Raton, FL 33431 USA. {\tt\small \{ahasankhani2019, tangy\}@fau.edu.}}
\thanks{E. B. Ondes and C. Sultan are with the Department of Aerospace and Ocean Engineering, Virginia Tech, 460 Old Turner St., Blacksburg, VA 24061, USA. {\tt\small \{ondes, csultan\}@vt.edu.}}%
\thanks{J. VanZwieten is with the Department of Civil, Environmental, and Geomatics Engineering, Florida Atlantic University, Boca Raton, FL 33431, USA. {\tt\small jvanzwi@fau.edu.}}
       }

\begin{document}

\maketitle
\thispagestyle{empty}
\pagestyle{empty}

\begin{abstract}
This paper presents an integrated path planning and tracking control of marine hydrokinetic energy harvesting devices. To address the highly nonlinear and uncertain oceanic environment, the path planner is designed based on a reinforcement learning (RL) approach by fully exploring the historical ocean current profiles. The planner will search for a path to optimize a chosen cost criterion, such as maximizing the total harvested energy for a given time. Model predictive control (MPC) is then utilized to design the tracking control for the optimal path command from the planner subject to problem constraints. The planner and the tracking control are accommodated in an integrated framework to optimize these two parts in a real-time manner. The proposed approach is validated on a marine current turbine (MCT) that executes vertical waypoint path searching to maximize the net power due to spatiotemporal uncertainties in the ocean environment, as well as the path following via an MPC tracking controller to navigate the MCT to the optimal path. Results demonstrate that the path planning increases harnessed power compared to the baseline (i.e., maintaining MCT at an equilibrium depth), and the tracking controller can successfully follow the reference path under different shear profiles.
\end{abstract}

\section{INTRODUCTION}
Research in marine hydrokinetic energy harvesting devices, including wave energy converter (WEC) and marine current turbine (MCT), is motivated today in the academic community and industry \cite{zhou2017developments}. Similar to an autonomous underwater vehicle (AUV), the MCT should autonomously navigate along the optimal path in the presence of a highly nonlinear and uncertain oceanic environment, accounting for the fact that the predicted ocean velocity is affected by perturbations.
An effective real-time path control may entail several prerequisites, leading to a challenging problem. The trajectory control should track the optimal path to minimize the tracking error, which should ideally converge to zero. Meanwhile, the optimal path is acquired by optimizing an ultimate goal, i.e., maximizing harvested energy from an MCT in a given time period. Moreover, the optimal path may change continuously due to the inherent spatiotemporal uncertainties in the oceanic environment \cite{lermusiaux2006quantifying}. Finally, the highly nonlinear dynamic model of the MCT that includes multi-physics couplings increases the problem complexity. Therefore, the challenges of real-time optimal path control is significant for this system.

The real-time path control is interpreted with a prohibitive large-scale control program, involving a complex dynamic model of the system, discrete inputs, discrete outputs that enforce waypoint tracking constraints while iteratively optimizing a chosen cost criterion via path planning. The path tracking problem has been addressed for a wide range of applications, including robotics \cite{kanayama1990stable,verscheure2009time}, unmanned aerial vehicles \cite{wang2017non,oliveira2016moving}, helicopters \cite{greer2020shrinking}, autonomous underwater vehicles \cite{encarnacao20003d}, and ocean current turbine \cite{ngo2017model}. For effective real-time path control design, the first task is to ensure that an optimal path is computed by the planner for the tracking controller as a reference path.

An integrated framework for planning and tracking control usually includes an upper-level planner to generate an optimal path and a lower-level tracker that is responsible for tracking the generated optimal path. 
For examples, the path planning and tracking control have been successfully proposed for airborne wind energy systems \cite{cobb2019flexible,todeschini2019control,schmidt2019flight,zgraggen2014real,koenemann2017modeling,vermillion2021electricity}, discrete manufacturing plants \cite{fagiano2020hierarchical}, autonomous vehicle \cite{ji2016path,maitland2018towards}, unmanned aerial vehicle \cite{pant2020co,jafari2018brain}, and autonomous underwater vehicle \cite{wang2020path,repoulias2007planar}. In similar research, the trajectory planning and control have been solved for a buoyancy controlled device for an underwater vehicle to achieve its vertical maneuvering \cite{zuo2019optimal}. However, to be computationally tractable, the system model and tracking controller have been significantly simplified.

In this paper, to deal with the real-time path control for marine energy harvesting devices, we propose an integrated approach that iteratively plans the optimal path to maximize the harvested power and tracks the commanded path to minimize the tracking error. The motivation is clear -- in order to maximize the energy harvesting of the device, the system vertical path should be real-time planned and tracked in the dynamic oceanic environment. Moreover, an integrated framework is intended as a simpler, computationally cheaper alternative to simultaneous path planning and tracking. Specifically, we develop a path planner that allocates an optimal path to maximize the harvested energy from an MCT using a deep reinforcement learning (DRL) approach. Model predictive control (MPC) is then developed to track the optimal reference path commanded from the planner subject to the MCT dynamic model and problem constraints. The ultimate goal of the tracking control is to safely navigate the turbine along the optimal path, considering the sluggish dynamics while avoiding any aggressive motion and instability. 


The remainder of this paper is organized as follows. Section~\ref{sec:prob. descrip} describes the modeling and overall structure of the integrated path planning and tracking control for the MCT. Section~\ref{sec:OCT} presents detailed path planning and path tracking. Section~\ref{sec:results} shows the numerical results. Finally, Section~\ref{sec:conclusion} concludes the paper.

\section{Problem Description} \label{sec:prob. descrip}
In this section, the overall problem is formulated for an application of the MCT, discussing the model of the turbine, oceanic uncertain environment, and the proposed path planning and control architecture for the MCT system.

\subsection{MCT Modeling} \label{sec:OCT_model}
\begin{figure}[t]
\centering
\includegraphics[width=0.92\linewidth]{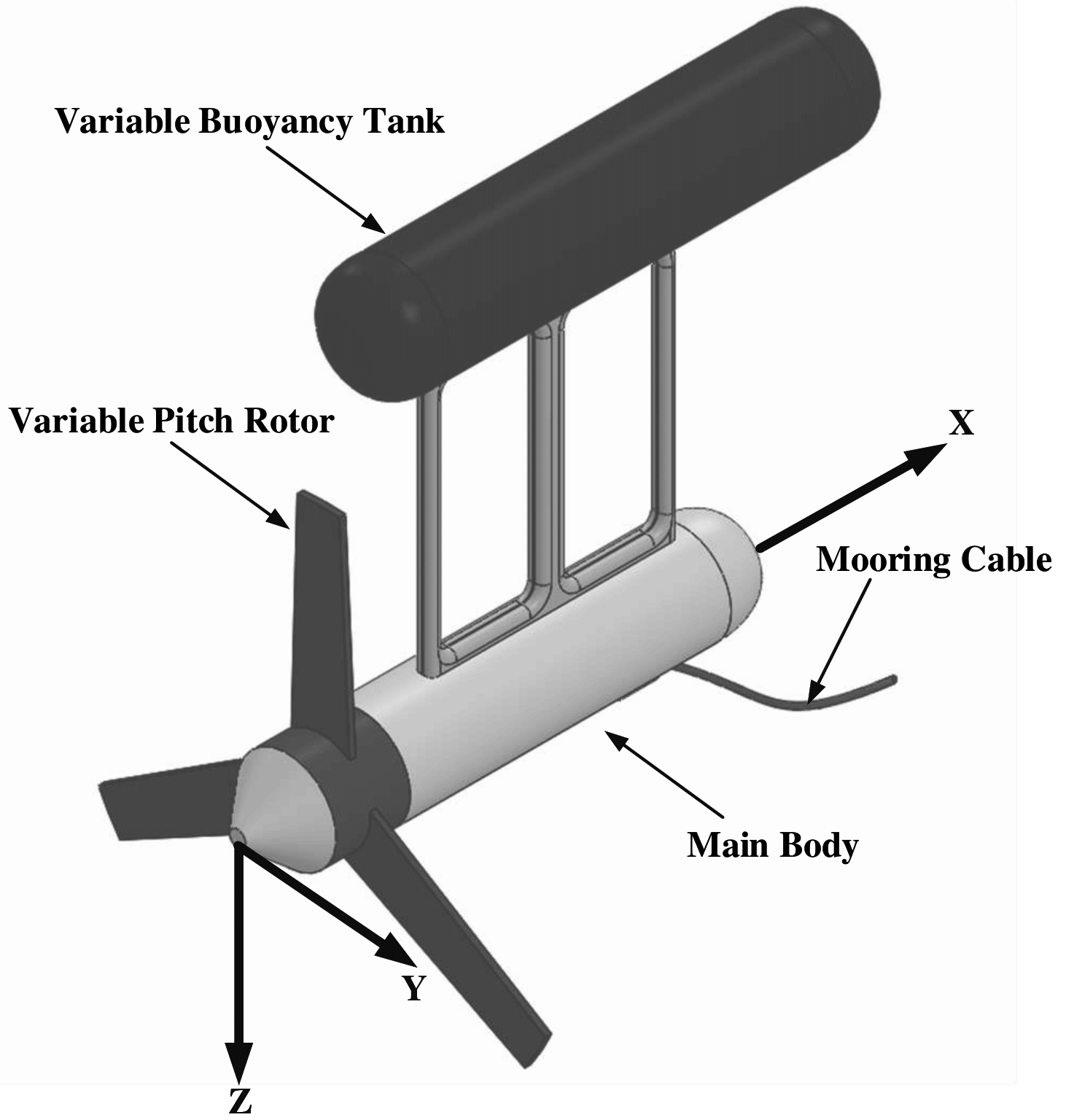}
\caption{Schematic diagram of the numerically simulated buoyancy controlled MCT in this paper \cite{hasankhani2021modeling}.}
\label{fig:OCT}
\vspace{-3mm}
\end{figure}

The investigated 700 $\,kW$ MCT in this paper (see Fig. \ref{fig:OCT}) is a representative design for operation in the Gulf Stream off Florida's East Coast, which consists of two variable buoyancy sections of a single variable buoyancy tank, variable pitch rotor, main body, and a 607 $\,m$ mooring cable attached to the ocean floor at a depth of 325 $\,m$ \cite{hasankhani2021modeling,vanzwieten2012numerical}. The designed MCT system and its parameter roughly follow the prototype systems from IHI Corp. \cite{ueno2018development} and the University of Naples \cite{coiro2017development}; still, the MCT includes a single rotor, where the tanks designed to operate at a depth of 50 $\,m$ in case of half-filled with ballast water at the ocean current speed of 1.6 $\,m/s$. A seven degrees-of-freedom (DOF) model is used to characterize the nonlinear dynamics of the MCT.


\textbf{Kinematics and Coordinate Frame:} In order to obtain the kinematics, the coordinate frames are defined for the MCT system: the inertial coordinate frame $(\mathcal{T}_I)$, the body-fixed coordinate frame $(\mathcal{T}_B)$, the momentum mesh coordinate frame $(\mathcal{T}_M)$, the shaft coordinate frame $(\mathcal{T}_S)$, and the rotor blade coordinate frame $(\mathcal{T}_R)$. $\mathcal{T}_I$ is situated at mean ocean level, with its X-axis pointing north, the Y-axis pointing east, and the Z-axis pointing downstream; $\mathcal{T}_B$ is placed at the MCT's center of gravity, with its X-axis pointing to the nose, the Z-axis pointing the bottom of MCT, and the Y-axis obtained through the right hand rule; $\mathcal{T}_M$ is situated at the center of rotor's rotation, with the axial direction parallel to the X-axis, the tangential direction pointing toward the rotor's rotational direction, and the radial direction pointing outward from the rotor's center; Finally, $\mathcal{T}_S$ is located at the center of rotor's rotation, and $\mathcal{T}_R$ is situated at the quarter cord line of rotor blade section \cite{vanzwieten2012numerical}. The transformation matrix from $\mathcal{T}_I$ to $\mathcal{T}_B$ is defined due to rotations about the yaw angle $\psi$, the pitch angle  $\theta$, and the roll angle $\phi$:
\begin{equation}\label{transformation matrix}
    T^{\mathcal{T}_B}_{\mathcal{T}_I}=
    \begin{bmatrix}
 c_\psi c_\theta &  s_\psi c_\theta & -s_\theta\\ 
 c_\psi s_\theta s_\phi-s_\psi c_\phi & c_\psi c_\phi+s_\psi s_\theta s_\phi & c_\theta s_\phi\\ 
 c_\psi s_\theta c_\phi-s_\psi s_\phi & -c_\psi s_\phi+s_\psi s_\theta c_\phi & c_\theta c_\phi\\
\end{bmatrix}
\end{equation}
where $s_{(.)}=sin(.)$ and $c_{(.)}=cos(.)$.

\textbf{Equations of Motion:} The motion of the buoyancy controlled MCT is modeled through the 7-DOF, consisting of 6-DOF motion of the main body and 1-DOF of rotor's rotation about the X-axis \cite{vanzwieten2012numerical}, namely:
\begin{equation}\label{equation of motion_1}
\resizebox{0.43\textwidth}{!}{$
    f_x= m^{v}(\dot{u}-vr+w q)-m^{v}x_{cg}(q^2+r^2)+m^{v}_{b}z^{v}_{cg_{b}}(p_b r+\dot{q})
$}
\end{equation}
\begin{equation}\label{equation of motion_2}
\resizebox{0.43\textwidth}{!}{$
\begin{split}
    f_y=& m^{v} (\dot{v}+ur)-w(m^{v}_{b}p_{b}+m^{v}_{r}p_{r})+ m^{v}_{b}z^{v}_{cg_{b}}(q_r-\dot{p}_b)\\
    &+m^{v}_{b}x^{v}_{cg_{b}} q p_b+m_r^{v} x_{cg_{r}}^{v} q p_r + m^{v} x^{v}_{cg} \dot{r}
    \end{split}
$}
\end{equation}
\begin{equation}\label{equation of motion_3}
\resizebox{0.43\textwidth}{!}{$
\begin{split}
    f_z=& m^{v} (\dot{w}-uq)+v(m^{v}_{b}p_{b}+m^{v}_{r}p_{r})+ m^{v}_{b}z^{v}_{cg_{b}}(p^{2}_{b}+q^2)\\
    &+m^{v}_{b}x^{v}_{cg_{b}} r p_b+m_r^{v} x_{cg_{r}}^{v} r p_r + m^{v} x^{v}_{cg} \dot{q}
    \end{split}
$}
\end{equation}
\begin{equation}\label{equation of motion_4}
\resizebox{0.43\textwidth}{!}{$
    M_{x_{b}}+\tau_{em}=-m^{v}_{b}z_{cg_{b}}^{v}(\dot{v}-w p_{b}+ur)+I^{v}_{x_{r}}\dot{p_{r}}+qr(I^{v}_{z_{r}}-I^{v}_{y_{r}})
    $}
\end{equation}
\begin{equation}\label{equation of motion_5}
\resizebox{0.43\textwidth}{!}{$
\begin{split}
    M_y=&I^{v}_{y} \dot{q}+ rp_{b}(I^{v}_{x_{b}}-I^{v}_{z_{b}}) + r p_{r} (I^{v}_{x_{r}}-I^{v}_{z_{r}})+I^{v}_{xz_{b}}(p_{b}^{2}-r^2)\\
    &+ m^{v}_{b} z^{v}_{cg_{b}}(\dot{u}-vr+w q)-m^{v} x^{v}_{cg}(\dot{w}-uq)+ m^{v}_{b} x^{v}_{cg_{b}} v p_{b}\\
    & - m_{r}^{v} x_{cg_{r}}^{v} v p_{r}
    \end{split}
    $}
\end{equation}
\begin{equation}\label{equation of motion_6}
\resizebox{0.43\textwidth}{!}{$
\begin{split}
    M_z=&I^{v}_{z} \dot{r}+ qp_{b}(I^{v}_{y_{b}}-I^{v}_{x_{b}}) + q p_{r} (I^{v}_{y_{r}}-I^{v}_{x_{r}})+I^{v}_{xz_{b}}(rq-\dot{p}_{b})\\
    &+ m^{v} x^{v}_{cg}(\dot{v}+ur)- m^{v}_{b} x^{v}_{cg_{b}} w p_{b} - m_{r}^{v} x_{cg_{r}}^{v} w p_{r}
    \end{split}
    $}
\end{equation}
\begin{equation}\label{equation of motion_7}
    \dot{p}_{r}=\frac{M_{x_{r}}-M_{x_{s}}-qr(I^{v}_{z_{r}}-I^{v}_{y_{r}})}{I^{v}_{x_r}}
\end{equation}
where $m$ is the mass; $u,v,w$ are the linear velocities of the MCT main body in $\mathcal{T}_{B}$; $p_b,q,r$ are the angular velocities of the MCT main body in $\mathcal{T}_{B}$; $p_{r}$ denotes angular velocity of the MCT rotor in $\mathcal{T}_{B}$; $x,y,z$ are location of the origin of the MCT main body; $\phi,\theta,\psi$ are the Euler angles of the MCT main body; $f_{(.)}$ is the total external force in the direction of $(.)$; $M_{(.)}$ is the total external moment in the direction of $(.)$; $I_{(.)}$ denotes the mass moment; $(.)^{v}$ denotes the virtual; $\tau_{em}$ is the electromechanical shaft torque; $(.)_{cg}$ denotes the center of gravity; Finally, $(.)_{r}$ and $(.)_{b}$ denote the rotor and the main body, respectively.

The total external load $f$, which is shown about $x-$, $y-$, and $z-$ axes in the equations of motion, is the sum of gravitational and buoyancy force $f_{gb}$, rotor force $f_r$, main body forces $f_b$, and cable force $f_c$, i.e. $f=f_{gb}+f_r+f_{b}+f_{c}$. Cable forces are calculated using a finite element lumped mass cable model and rotor forces are calculated using a blade element momentum based rotor model, with a dynamic wake inflow algorithm (see \cite{vanzwieten2012numerical} for details).

\textbf{Linear Model:} The linearization process averages the MCT dynamics over the rotor rotation to remove the dependence on rotor azimuth angle and cable node states as suggested in \cite{ngo2020constrained} and is expanded to account for variable buoyancy control as proposed in \cite{Hasankhani_spatiotemporal} (with the justification that the linear and nonlinear models are in good agreement). 
The nonlinear dynamic model of the MCT is linearized around the equilibrium points, namely:
\begin{equation}\label{linear model}
\begin{split}
	\Delta \dot{x} &= A \Delta x + B \Delta u\\
	y &= C \Delta x 
	\end{split}
\end{equation}
where $A=\frac{\partial f}{\partial x}|^{x=x_{eq}}_{u=u_{eq}}$, $B=\frac{\partial f}{\partial u}|^{x=x_{eq}}_{u=u_{eq}}$, $C=\frac{\partial g}{\partial u}|^{x=x_{eq}}_{u=u_{eq}}$,  $\Delta(.)\overset{\Delta}{=}(.)-(.)_{eq}$, with $eq$ denoting the equilibrium values. In order to maintain the simplicity, $\Delta$ is not shown for the remainder of the paper. $x\in \mathbb{R}^{13}$, $u \in \mathbb{R}^3$, and $y \in \mathbb{R}$ are defined as $ x=[u~ v~ w~ p_b~ p_r~ q~ r~ x~ y~ z~ \phi~ \theta~ \psi]$, $u=[ B_f~ B_a~ \tau_{em}]$, and $y=[z]$. The control inputs are determined by two buoyancy tank fill fractions $B_f$ and $B_a$ denoting the fill fractions of the forward tank and the fill fraction of the aft tank, respectively, introducing into equations of motion by gravitational and buoyancy force $f_{gb}$.  

Furthermore, since the real-time path controller is designed in discrete-time with future objectives of implementation in a real plant (e.g., an MCT-based power plant), the linearized model is discretized using a sampling time, $T_s$, determined by the Nyquist-Shannon's sampling theorem \cite{1455040}. The corresponding discrete linear time-invariant (DLTI) plant model at discrete-time instant $k$ is:
\begin{equation}
\begin{split}
	x(k+1) &= A_{d} x(k) + B_{d} u(k)\\
    y(k) &= C_{d} x(k) 
\label{linea_model_discrete}
\end{split}
\end{equation}
with equilibrium points of $x_{eq}=[0~ 0~ 0~ 0~ 1.49~ 0~ 0~ 554.50~ \\0.38~ 50~ 0.01~ 0.00~ 3.14]$ and $u_{eq}=[0.4677~ 0.4677~ -188280]$.

\subsection{Ocean Environment Modeling} \label{sec:Ocean_model}
\begin{figure}[t]
    \begin{center}
    \includegraphics[width=0.92\linewidth]{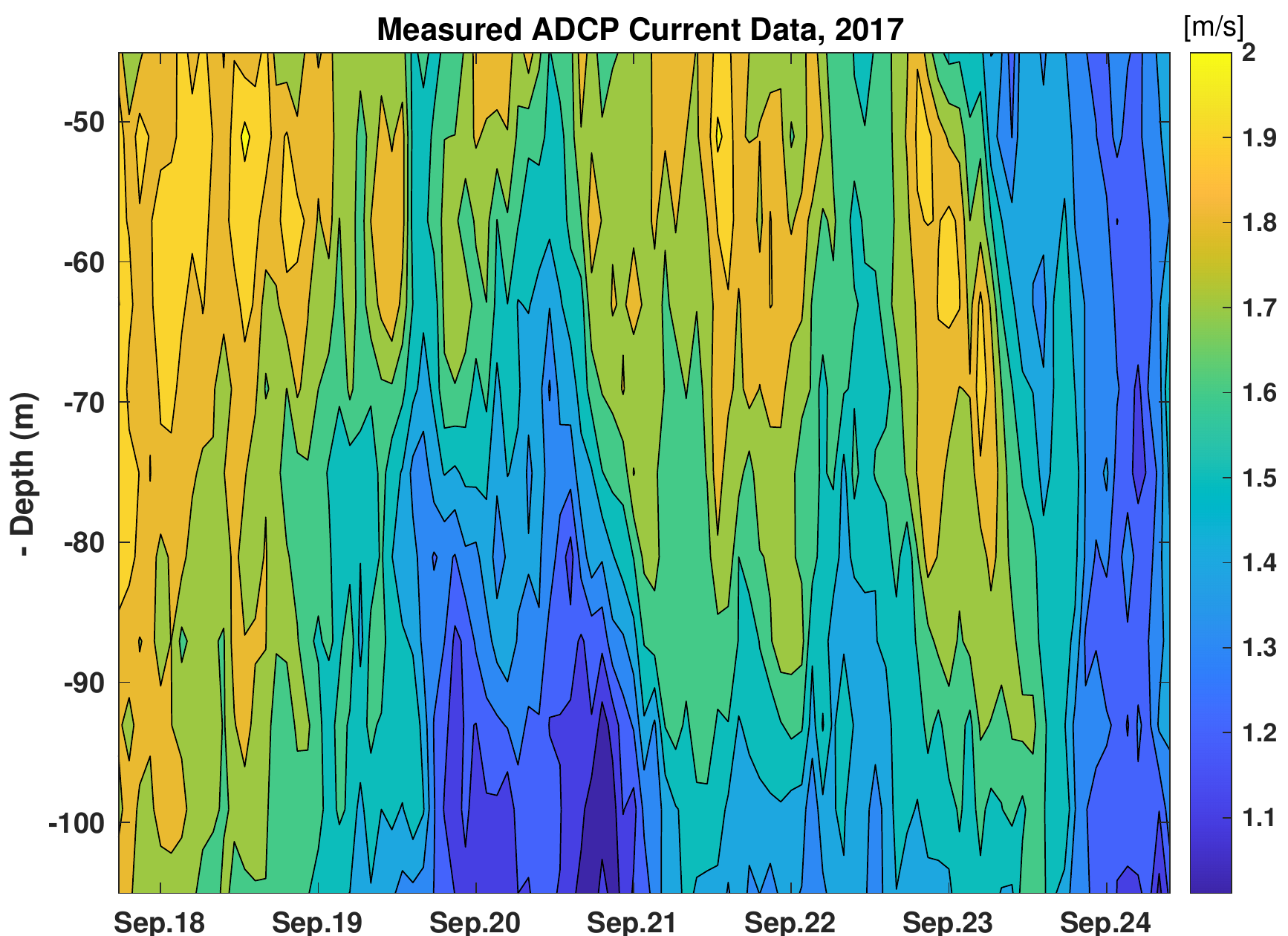}
    \end{center}
    \caption{Time histories of the current flow recorded by a 75 kHz ADCP at a latitude of $26.09 ^{\circ}N$ and longitude of $-79.80 ^{\circ}E$ represented for a sample one-week period.}
    \label{fig:velocity}
   \vspace{-3mm}
\end{figure}

To model the spatiotemporal uncertainties in the ocean environment, it is important to use real data. The spatial and temporal fluctuations in the current flow are resulted from turbulence, waves, and lower frequency flow structures. In this paper, we use the field measured data by a 75 kHz acoustic Doppler current profiler (ADCP) presented in \cite{Machado}, which were recorded at a latitude of $26.09 ^{\circ}N$ and longitude of $-79.80 ^{\circ}E$, as shown in Fig. \ref{fig:velocity}. The measurement resolution was $5 \,m$ within 400 $\,m$ water depth, where various current flow data (i.e., current speed, northward current velocity, eastward current velocity, etc.) were recorded. These measured current data were filtered to remove bad data as described in \cite{Machado}, which were mostly measured above a depth of 50 $\,m$.

MCT will operate spatiotemporally in this dynamic environment to harvest the ocean current energy. To address its real-time path planning and tracking, it is necessary to model and predict the uncertain ocean environment using the recorded ADCP data. In this paper, we leverage \emph{Gaussian process (GP)} modeling to characterize a statistical model of the environment. The GP models an input vector of spatiotemporal uncertain environment $\eta$ by $e=f(\eta)+\epsilon$, where $e$ denotes the observations from the environment, and $\epsilon\sim\mathcal{N}(0,\sigma^2)$ specifies the Gaussian noise:
\begin{equation}\label{GP}
    f(\eta)\sim \mathcal{N}(\mathcal{M}(\eta),\mathcal{K}(\eta,\eta^{'}))
\end{equation}
\begin{equation}\label{GP_mean}
   \mathcal{M}(\eta)=\mathbb{E}[f(\eta)]
\end{equation}
\begin{equation}\label{GP_covariance}
    \mathcal{K}(\eta,\eta^{'})=\mathbb{E}[(f(\eta)-\mathcal{M}(\eta))(f(\eta^{'})-\mathcal{M}(\eta^{'}))]
\end{equation}
where $\mathcal{M}(\eta)$ is the mean function, and $\mathcal{K}(\eta,\eta^{'})$ is the covariance function between $\eta$ and new enviornmental data denoted as $\eta^{'}$. Note that GP modeling ensures an \emph{accurate} model in the presence of a \emph{sufficiently rich} database \cite{ma2017informative}, denoted by $D=[(\eta,e)|\eta\in H, e\in E]$, from historical ocean current profile observations.

\subsection{Proposed Integrated Path Planning and Tracking Control}
To address the coupled MCT path planning/tracking control problem, we develop an \emph{integrated control} framework. The goal of the proposed architecture is to plan a cost criterion-optimal path and to track this optimal path, generally formulated in discrete-time in the following. To ensure an admissible path while avoiding the computational complexity, the path planning is constrained with the major operational constraints of the MCT instead of considering the detailed plant dynamics, which is considered in the path tracking.

\noindent \underline{\textit{Path Planning:}}
\begin{equation}\label{OF_planning}
\resizebox{0.43\textwidth}{!}{$
\begin{split}
J_{p}(x(.),u(.),y(.)) = \underset{y(.)}{\max}&[ l_{f}(x(N_{p}|k),u(N_{p}|k),y(N_{p}|k))\\
        & +\sum_{i=k}^{k+N_{p}-1}l(x(i|k),u(i|k)),y(i|k)]\\
\end{split}
$}
\end{equation}
s.t.
\begin{equation}\label{constraint1}
\resizebox{0.43\textwidth}{!}{$
e(i+1|k)\sim \mathcal{N}(\mathcal{M}(\eta),\mathcal{K}(\eta,\eta^{'}),~\forall i\in[0:N_{p}-1]
$}
\end{equation}
\begin{equation}\label{constraint2}
    h_{ip}(x(i),u(i),y(i))\leq0
\end{equation}
\begin{equation}\label{constraint3}
    h_{ep}(x(i),u(i),y(i))=0
\end{equation}

\noindent \underline{\textit{Path Tracking:}}
\begin{equation}\label{OF_tracking}
\begin{split}
J_{t}(x(.),u(.),y(.)) = & \underset{u(.)}{\min}[|y(N_{t}|k)-y^{*}(N_{t}|k)|^2\\ & + \sum_{i=k}^{k+N_{t}-1}|y(i|k)-y^{*}(i|k)|^2]
    \end{split}
\end{equation}
s.t.
\begin{align}\label{constraint4}
x(i+1|k) &= Ax(i|k) + Bu(i|k),~\forall i\in[0:N_{t}-1]
\end{align}
\begin{equation}\label{constraint5}
    h_{it}(x(i),u(i),y(i))\leq0
\end{equation}
\begin{equation}\label{constraint6}
    h_{et}(x(i),u(i),y(i))=0
\end{equation}
where $l\in\mathbb{R}$ and $l_{f}\in\mathbb{R}$ are the cost criterion and the terminal cost, respectively. Constraint (\ref{constraint1}) represents the uncertain environment profile, where $e\in\mathbb{R}^o$ is the environment data vector, $h_{i.}$ and $h_{e.}$ are inequality and equality constraint functions, respectively. $y$ and $y^{*}$ denotes the true waypoint path and optimal waypoint path (reference waypoint path for path tracking) for the MCT, $N_{p}$ and $N_{t}$ denote the prediction horizon for path planning and prediction horizon for path tracking, respectively. Constraint (\ref{constraint4}) represents the linearized model of the plant. The notation $x(i|k)$ represents the prediction of what the value of $x$ will be at time $i$ given the value of $x$ at time $k$.

\begin{figure}[t]
\centering
\includegraphics[width=0.95\linewidth]{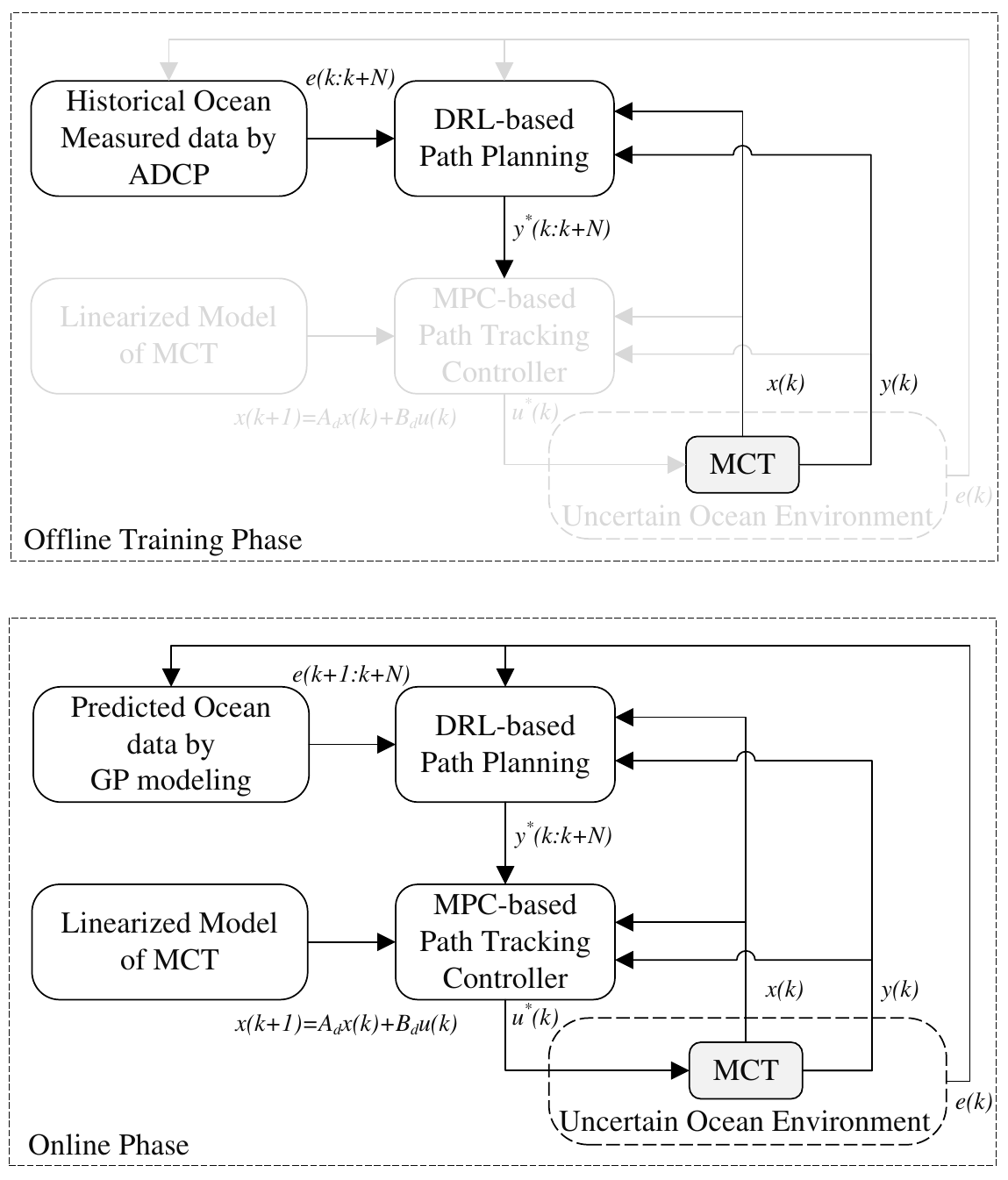}
\caption{Proposed integrated path planning and tracking control framework for marine current turbine.} \label{fig:control co-design}
\end{figure}

Fig. \ref{fig:control co-design} visualizes the proposed integrated framework, which contains two phases. In the offline phase, the planner is trained due to the historical ocean environment profile to find the optimal path. In the online phase, at each iteration, the path planner takes the previous iteration's state, $x(k)$, and environmental data, $e(k)$, to generate an optimal path $y^{*}(k)$. At each discrete-time instant, $k$, the path tracking controller follows the commanded optimal path. To be more specific, the path planner solves the optimization problem (\ref{OF_planning}) to maximize a user-defined cost criterion, where the optimal path satisfies the constraints corresponded to the uncertain ocean model (\ref{constraint1}). A learning-based approach will be utilized to find the optimal path through learning from the historical environment profile (details in Section \ref{sec:path planning}). The path tracker follows the optimal path assigned by the path planner and computes feasible control inputs, $u(k)$, to minimize the tracking error (\ref{OF_tracking}) by taking into account the linearized model of the plant (\ref{constraint4}) subject to the system constraints. An MPC-based approach will be used to solve the optimization problem as a quadratic programming problem using the linearized MCT model (details in Section \ref{sec:path tracking}).

\begin{remark}
The imposition of a learning-based approach, DRL, for the upper-level path planning is significant to address the highly uncertain environment through learning from the historical ocean environment profile. On the other hand, in the presence of the dynamic model of the plant, the selection of an analytical approach, MPC, for the lower-level path tracking ensures to successfully follow the assigned optimal path robustly, without violating the system's limitations. The result of this framework is an integrated control approach that satisfies path planning in an uncertain environment and path tracking using a detailed plant model.
\end{remark}

\section{Detailed Path planning and Path tracking} \label{sec:OCT}

\subsection{Path Planning through Reinforcement Learning} \label{sec:path planning}
The objective of path planning is to optimize a chosen cost criterion, considering the plant in a dynamic and highly uncertain environment. Specifically, for the MCT, the major objective is to maximize the harvested power from the plant. To solve the real-time path planning, denoted as an iterative decision making (control) problem in a stochastic environment, a DRL approach is considered as a powerful tool to find the optimal path through learning from the historical data, which shows acceptable robustness to the environmental uncertainties in similar applications \cite{lin2020comparison}. A brief review of the RL preliminaries is presented below.

\textbf{Reinforcement Learning (RL):} In RL, a Markov decision process (MDP), as a canonical formalism for the stochastic control problem, is modeled by a tuple $\mathcal{MDP}:=<\mathcal{S},\mathcal{A},\mathcal{P},R,\gamma>$, where $\mathcal{S}$ is the state space, $\mathcal{A}$ is the action space, $\mathcal{P}$ is the state transition probability, with $\mathcal{P}(s'|s,a)$ denoting the probability of transiting to the state $s'\in \mathcal{S}$ given the current state $s\in \mathcal{S}$ after taking the action $a\in \mathcal{A}$, $R:\mathcal{S} \times \mathcal{A}\rightarrow \mathbb{R}$ is the reward function, and $\gamma \in [0,1)$ is the discount factor. 

A policy $\pi:\mathcal{S}\rightarrow \mathcal{A}$ determines the action to take at state $s$. To estimate how good a particular policy will be with a given state, a value function $V_{\pi}(s)$ is used to measure the accumulated reward, which is defined by:
\begin{equation}\label{Value func}
V_{\pi}(s)=\sum_{a\in \mathcal{A}}\pi(a|s)\sum_{s'\in \mathcal{S}}\mathcal{P}(s'|s,a)(r_{k}+\gamma V_{\pi}(s'))
\end{equation}
where $\pi(a|s)$ denotes a policy in the RL to take an action $a\in\mathcal{A}$ at state $s\in\mathcal{S}$ and $r_{k}=R(s_k,a_k)$. Similarly, a Q-value function of a policy $\pi$ is defined as:
\begin{equation}\label{Q-Value func}
Q_{\pi}(s,a)=r_k+\gamma\sum_{s'\in \mathcal{S}}\mathcal{P}(s'|s,a)V_{\pi}(s')
\end{equation}

The optimal Q-value function $Q^{*}$ is defined as $Q^{*}(s,a)=\underset{\pi}\max Q_{\pi}(s,a)$. Given $Q^{*}$, the optimal policy $\pi^{*}$ is then calculated by:
\begin{equation}\label{optimal policy}
\pi^{*}=\underset{\pi}{argmax} \; Q_{\pi}(s,a)
\end{equation}

\emph{Q-learning} approach, as one of a popular RL algorithms, is chosen to find $Q^{*}(s,a)$; however, a standard tabular Q-learning lacks to scale to problems with high-dimensional state and action spaces. To tame this issue, deep Q-learning is introduced to approximate the Q-value function for all possible actions with a deep neural network. To find the optimal Q-value $Q^{*}$, we use the DRL approach (specifically deep Q-network (DQN)). The DQN includes two deep neural networks named the \emph{Q-network} $Q(.;\theta)$ and \emph{target network} $Q(.;\theta^{-})$, where $\theta$ and $\theta^{-}$ denote the parameters (weights) of these two networks. The main goal of the DQN is to minimize the distance between $Q(s,a)$ estimated by $Q(.;\theta)$ and temporal difference target $t_j$ estimated by the $\widehat{Q}(.;\theta^{-})$, expressed as a gradient descent:
\begin{equation}\label{gradient descent}
L(\theta)=[t_{j}-Q(s_j,a_j;\theta)]^2
\end{equation}
where
\begin{equation}\label{TD}
t_j:=r_j+\gamma\underset{a'}{\max}Q(s_{j+1},a';\theta^{-}))
\end{equation}

To select an action, we follow an adaptive $\epsilon-$greedy policy \cite{liang2019deep} as a method to choose random actions with uniform distribution from possible actions, namely:
\begin{equation}\label{epsilon greedy}
    a_k=
\begin{cases}
    arg\underset{a_k}\max \: Q(s_k,a_k), &   1-\epsilon\\
    \text{random}~a\in A, & \epsilon
\end{cases}\end{equation}
\begin{equation}\label{epsilon}
    \epsilon=\epsilon_{min}+(\epsilon_{max}-\epsilon_{min})e^{-d\times n_{e}}
\end{equation}
where $d$ is the decay factor and $n_{e}$ is the episode number. 

\textbf{DRL-based Path Planning:}   
For the MCT application, at each iteration, the DRL-based approach decides on whether to keep the MCT at its current position (i.e., water depth) or to change the position that has a large flow speed to maximize the harvested power in the planning prediction horizon $N_p$ (see (\ref{OF_planning})). The state $\mathcal{S}$, action $\mathcal{A}$, and reward $\mathcal{R}$ notations should be defined according to the MCT-specific application. The MCT can observe its current states, $x(k)$, and the ocean current velocity $e(k)\triangleq v_{c}(k)$; hence, the state of this system at time $k$ is $s(k)=[x(k),v_{c}(k)]$. To define the action space, we consider the possible action that MCT can take to maximize the generated power, where the position change constitutes the action of MCT at $k$ (i.e., $a(k)\overset{\Delta}{=}y(k)$). 

The major concern for the path planning problem for the MCT system arises from its complex power equation and how to define the DRL's reward in relation to that power expression. 
Since the ultimate objective of path planning is to maximize the generated power, a positive reward, $\alpha$, is gained if the position change fulfills the power increase, and a zero reward is given to the wrong position change:
\begin{equation}\label{reward_OCT}
    R(s_k,a_k)= 
\begin{cases}
    \alpha ,&  \text{$P(z(k+1))> P(z(k))$}\\
    0 ,&  \text{else}\\
\end{cases}\end{equation}

The total harvested power $P(z(k))$ from the MCT comprises of three terms of generated power $P^{G}$, consumed power to hold depth $P^{HD}$, and consumed power to change depth $P^{CD}$, namely:
\begin{equation}\label{Power}
    P(z(k)) = P^{G} - P^{HD} - P^{CD}
\end{equation}
where
\begin{equation}\label{Power_OCT}
    P^{G} = \frac{1}{2} \rho A c_{p} v_{c}^{3}(z(k))
\end{equation}
\begin{equation}\label{Pump Power HD}
\resizebox{0.43\textwidth}{!}{$
    P^{HD} =
    \begin{cases}
    0 ,&  \text{$v_{c}(z(k+1))-v_{c}(z(k)) < 0$}\\
    \frac{\zeta_{1}}{T_{p}}[v_{c}(z(k+1))-v_{c}(z(k))],&  \text{$v_{c}(z(k+1))-v_{c}(z(k)) > 0$}\\
\end{cases}
$}
\end{equation}
\begin{equation}\label{Pump Power CD}
\resizebox{0.43\textwidth}{!}{$
    P^{CD} =
    \begin{cases}
    0 ,&  \text{$z(k+1)-z(k) > 0$}\\
    \frac{\zeta_{2}}{T_{p}}[z(k+1)-z(k)],&  \text{$z(k+1)-z(k) < 0$}\\
\end{cases}
$}
\end{equation}
where $\rho$ denotes the water density, $A$ is the swept area by the MCT rotor, $v_{c}$ denotes the ocean velocity (denoted as $e$ in our framework presented in Fig. \ref{fig:control co-design}), $c_{p}$ is the average power coefﬁcient, $\zeta_{1}$ and $\zeta_{2}$ denote the coefficients for the power equations, and $T_{p}$ denotes the path planning sampling time.

According to the power equation (\ref{Power}), the optimal depth (denoted the optimal power) is a compromise between the maximum water current and both the power consumption to hold depth and power consumption to change the depth. On the other hand, note that the magnitude of a non-zero reward $\alpha$ significantly impacts the training procedure. We formulate the path planning problem to maximize the cumulative reward gained through the position change, namely:
\begin{equation}\label{path planning_OCT}
\pi^{*}=\underset{\pi}{argmax} \; [r_k+\gamma\sum_{s'\in \mathcal{S}}\mathcal{P}(s'|s,a)V_{\pi}(s')]
\end{equation}
where the MCT learns the optimal policy $\pi^{*}$ using DQN approach to find the optimal waypoint path by solving (\ref{path planning_OCT}).

To do the path planning, we should first construct the DQN through offline training using the historical ocean environmental data, and the constructed DQN is then used to find the optimal path in an online phase. Algorithm \ref{alg:DRL} details the offline training process.

\begin{algorithm}[t]
	\caption{Deep reinforcement learning for path planning}\label{alg:DRL}
	\begin{algorithmic}[1]
	    \State Initialize recorded training data sample of an uncertain environment;
	    \State Initialize replay memory $M$ and mini-batch size;
	    \State Initialize action-value function $Q$ with random weights $\theta$;
	    \State Initialize target action-value function $\widehat{Q}$ with random weights $\theta^{-}=\theta$;
		\For {episode $i=1 \; to \; N_\text{episode}$}
		\State Initialize initial state;
			\For {time step $k=1 \; to \; N_{p}$}
			   \State Select action $a_k$ using $\epsilon$-greedy policy (\ref{epsilon greedy});
			   \State Take action $a_k$ and observe $r_{k}$ and $s_{k+1}$;
			   \State Store transition $(s_k,a_k,r_k,s_{k+1})$ in $M$;
			   \State \multiline{%
			   Sample random mini-batch of transitions $(s_j,a_j,r_j,s_{j+1})$ from $M$;}
			   \State Set $t_j=r_j+\gamma\underset{a'}{\max}Q(s_{j+1},a';\theta^{-}))$;
			   \State \multiline{%
			   Perform a gradient descent step on $L(\theta)=[t_{j}-Q(s_j,a_j;\theta)]^2$ with respect to $\theta$;}
			   \State Update $\widehat{Q}$ every $b$ steps and set $\theta^{-}=\theta$;
			\EndFor
		\EndFor
		\State Output offline trained optimal deep Q-network $Q^{*}$;
	\end{algorithmic} 
\end{algorithm}

\begin{algorithm}[t]
	\caption{Model predictive control for path tracking}\label{alg:MPC}
	\begin{algorithmic}[1]
		\State Discretize the continuous linear model based on Shannon's Theorem;
		\State Initialize the linear model inside the MPC solver; 
		\State Initialize the constraints for the actuators and the system states;
		\State Initialize $N_t$, $N_c$, $Q_{\text{part}}$, $Q_\text{term}$, $R_{\Delta \text {part}}$ and $R_{\text {part}}$;
		\State Initialize the initial state and control inputs;
		\For {$i=1 \; to \; N_\text{operation}$}
		    \State \multiline{%
		    Initialize the reference trajectory based on the optimal waypoint path from the DRL;}
			\For {$j=1 \; to \; N_\text{operation-int}$}
				\State \multiline{%
				Apply the interior-point method, select $ u^* $ for $ N_c $ over $ N_t $;}
				\State \multiline{%
				Solve the cost function $ \tilde{V} $ in (\ref{OF_MPC}) and obtain optimal $[u^*(i|N_c),...,u^*(i|i+N_c-1)]$;}
				\State \multiline{%
				Apply $u^*(i|N_c)$ to the DLTI plant and obtain the next system state $x(i+1)$;
				}
			\EndFor
			\State \multiline{%
			Update the position error and send a feedback to DRL;
			}
		\EndFor
	\end{algorithmic}  
\end{algorithm}

\subsection{Path Tracking through Model Predictive Control} \label{sec:path tracking}
The path tracking will minimize the distance between the plant's current path and the optimal path commanded by the path planner. MPC, as a powerful model-based algorithm, is chosen to address path tracking due to the complex dynamics consist of multiple states and the availability of a detailed MCT model.

\textbf{Model Predictive Control (MPC):}
To design an MPC for the DLTI system defined in (\ref{linea_model_discrete}), the following cost function must be minimized:
\begin{equation}\label{OF_MPC}
\resizebox{0.43\textwidth}{!}{$
    \begin{split}
    	\tilde{V} \equiv & \frac{1}{2} 
    	\left(
    	    \sum_{i=k+1}^{k+N_{t}-1}
    	    \left(
    	        \left(y(i|k)-y^{*}(i)\right)^{T} 
    	        Q_\text{part}
    	        \left(y(i|k)-y^{*}(i)\right)
	        \right)
	    \right.\\
    	&   +\sum_{i=k}^{k+N_{c}-1}
    	    \left(
    	        \left(u(i | k)-u(i-1 | k)\right)^{T} 
    	        R_{\Delta \text {part}}
    	        \left(u(i | k)-u(i-1 | k)\right)
	        \right.\\
    	&       +\left.\left(u(i | k)-d(i)\right)^{T} 
    	        R_{\text {part}}
    	        \left(u(i | k)-d(i)\right)
    	    \right)\\
    	&\left.
    	    +\left(y(k+N_{t} | k)-y^{*}(k+N_{t})\right)^{T} 
    	    Q_{\text {term}}
    	    \left(y(k+N_{t} | k)-y^{*}(k+N_{t})\right)
    	\right)
    \end{split}
    $}
\end{equation}
subject to the constraints:
\begin{align}
	u^{\min }(i)        &\leq u(i) \leq u^{\max }(i) \label{const1_MPC}\\ 
	\Delta u^{\min }(i) &\leq \Delta u(i) \leq \Delta u^{\max }(i) \label{const2_MPC}\\ 
	y^{\min }(i)      &\leq y(i) \leq y^{\max}(i) \label{const3_MPC} 
\end{align}
where $N_{t}$ and $N_{c}$ are the prediction horizon and the control horizon, respectively; $ d(i) $ is a vector of the arbitrary user-selected target controls at time step $i$, $ Q_{\text{part}}$ and $ Q_\text{term}$ are the weight factors for the states and the final state point, respectively. Finally, $ R_{\Delta \text {part}} $ and $ R_{\text {part}} $ are the weight factors for the control input changes and the desired control inputs, respectively.
Constraints (\ref{const1_MPC})-(\ref{const2_MPC}) impose that control input and control input change should stay within a predefined bound. Constraint (\ref{const3_MPC}) captures the need to bound the system output. Here, $(.)^{\min }(i)$ and $(.)^{\max}(i)$ are the minimum and maximum values of $(.)(i)$. 
    
After the algebraic manipulations as given in \cite{greer2020shrinking}, the cost function is converted to the quadratic form, which is then minimized using the interior-point method, subject to the constraints given in (\ref{const1_MPC})-(\ref{const3_MPC}).



\textbf{MPC-based Path Tracking:} In our problem, we will leverage the path tracking algorithm to obligate the plant to follow the optimal path with a minimum error subject to system constraints. Considering the MCT with a high number of coupled states and the requirement of handling many constraints, MPC provides a controller that meets the desired objectives in a guaranteed manner. Also, the user-defined weights bring flexibility about choosing the more important states in the plant to be followed while giving less priority to the other states that are relatively less crucial for the desired performance. MPC is designed to minimize the cost function as given in (\ref{OF_MPC}), in which the weights are defined to prioritize the states to follow the trajectory with the minimum error. The reference path is defined in the vector of target outputs, $y^{*}(i)$, and at each sampling time, the updated reference is received from the DRL-based path planner. Algorithm \ref{alg:MPC} describes the MPC-based path tracking.

The constraints of the MPC solver are defined based on the MCT design and specifications. Control inputs saturations are defined as:
\begin{align}\label{constraint_controller input}
    {\left[\left|B_f\right|,\left|B_a\right|,\left|\tau_{e m}\right|\right] \leq\left[\% 50, \% 50 , 0.2 \times \tau_{e m}^{e q} \mathrm{Nm}\right]}
\end{align}
and the slew rates constraints are:
\begin{align}
	{\left[\left|\dot{B}_{\text{aft}}\right|,\left|\dot{B}_{\text{front}}\right|\right] \leq[7.45\times10^{-4},7.45\times10^{-4}] \% / \mathrm{s}}
\end{align}

Since the desired performance is to track the depth reference, the related state, $z$, has the only weight defined in $Q_\text{part}$ and the rest is set to be zero as $Q_\text{part} = \operatorname{diag}(0,0,0,
                    0,0,0,0,
                    0,0,1,
                    0,0,0)$.
The other rates are defined as \(Q_{\text {term }}= I_{13}, R_{\text{part}}=0,\) and \(R_{\Delta \text{part}}=I_{3},\) where \(I_{n}\) is the \(n \times n\) identity matrix. 

\begin{figure}[t]
\centering
\includegraphics[width=0.95\linewidth]{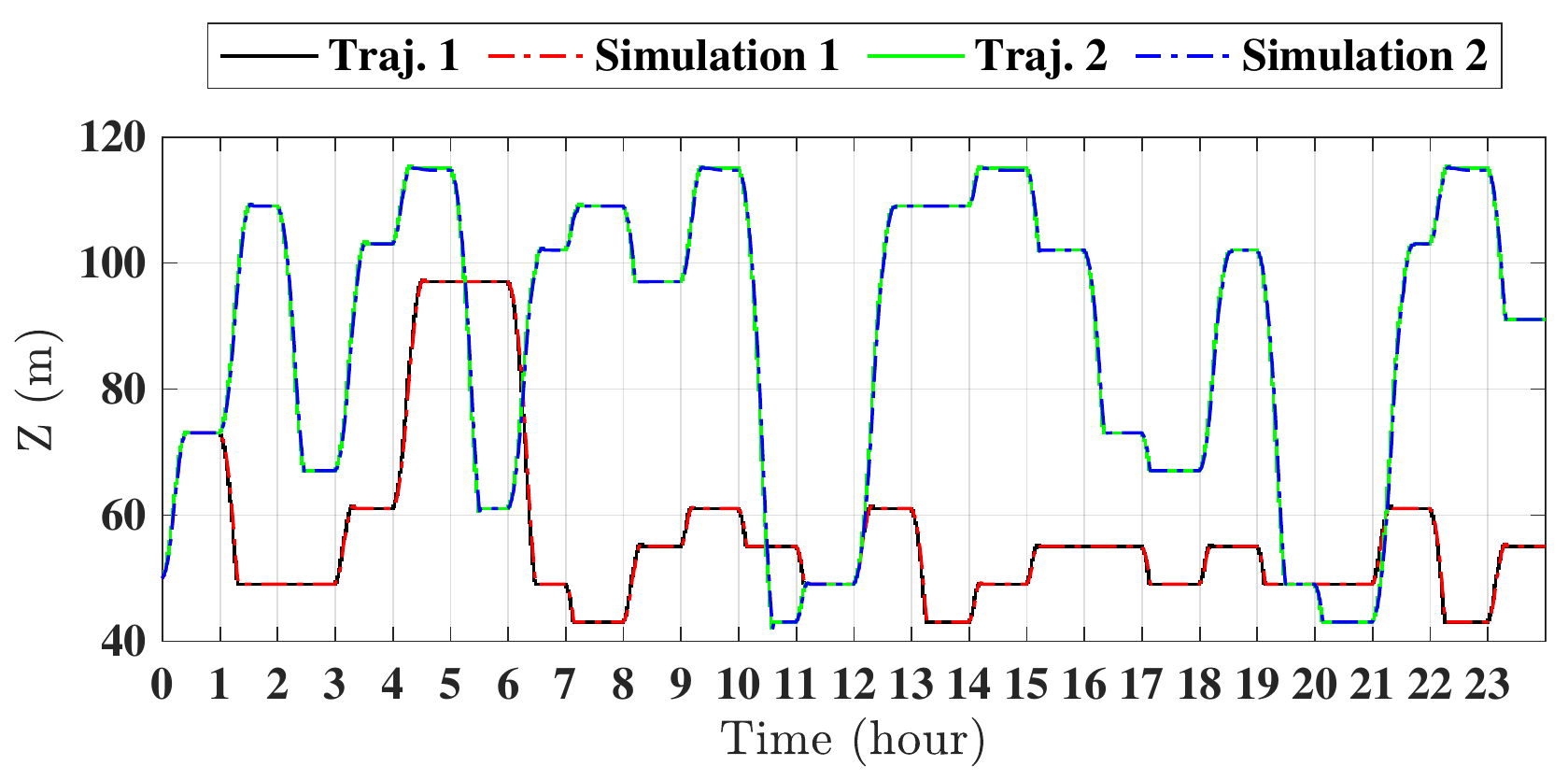}
\caption{Trajectory followed by the MCT along with the optimal reference trajectory for two ocean shear profiles.}
\label{fig:trajectory}
\end{figure}

\section{Numerical Results} \label{sec:results}
We evaluate the proposed path planning and path tracking for the MCT with the total mass of $4.98\times10^5$ $\,kg$, and mass moments of $I_{xx}=1.35\times10^7$ $\,kg m^3$, $I_{yy}=4.74\times10^7$ $\,kg m^3$, and $I_{zz}=3.45\times10^7$ $\,kg m^3$ \cite{hasankhani2021modeling}. The other parameters for the power equations include $\rho=1030 \,kg/m^3$, $A=100\pi$, $c_{p}=41.5 \%$, $\zeta_{1}=9.113$, and $\zeta_{2}=-0.0365$. For the path planner, we select a DQN with two hidden layers with a buffer size of $5e5$, a batch size of 64, $\gamma=0.5$,  $\epsilon_{min}=0.01$, and $\epsilon_{max}=1$. The experimental tests show that setting the reward value to $\alpha=+1$ 
justifies a successful training in our problem. For the tracking controller, the horizons are selected as $N_t=40$, and $N_c = 20$ considering the system's performance with highly-limited actuator slew rates. To find the sampling time, the eigenvector analysis is carried out and the eigenvalues of the linear MCT model are:
\begin{align}
    \begin{array}{cc}
    \begin{bmatrix}
        -0.2731 \pm 1.2585 i & -0.1121 \pm 0.1549 i \cr
        -0.2588 \pm 0.9618 i & -0.0033 \pm 0.0021 i \cr
        -0.2647 \pm 0.3564 i & -0.0483 \cr
        -0.1754 \pm 0.3793 i &
    \end{bmatrix}
    \end{array}
\end{align}

According to this analysis, the maximum frequency of the MCT system is calculated $\omega_{n,max}=1.29\,rad/s$ so that we opt for a sampling time of $T_s=2\,s$ satisfying the minimum requirement based on the Shannon sampling theorem. 

\begin{figure}[t]
\centering
\includegraphics[width=0.95\linewidth]{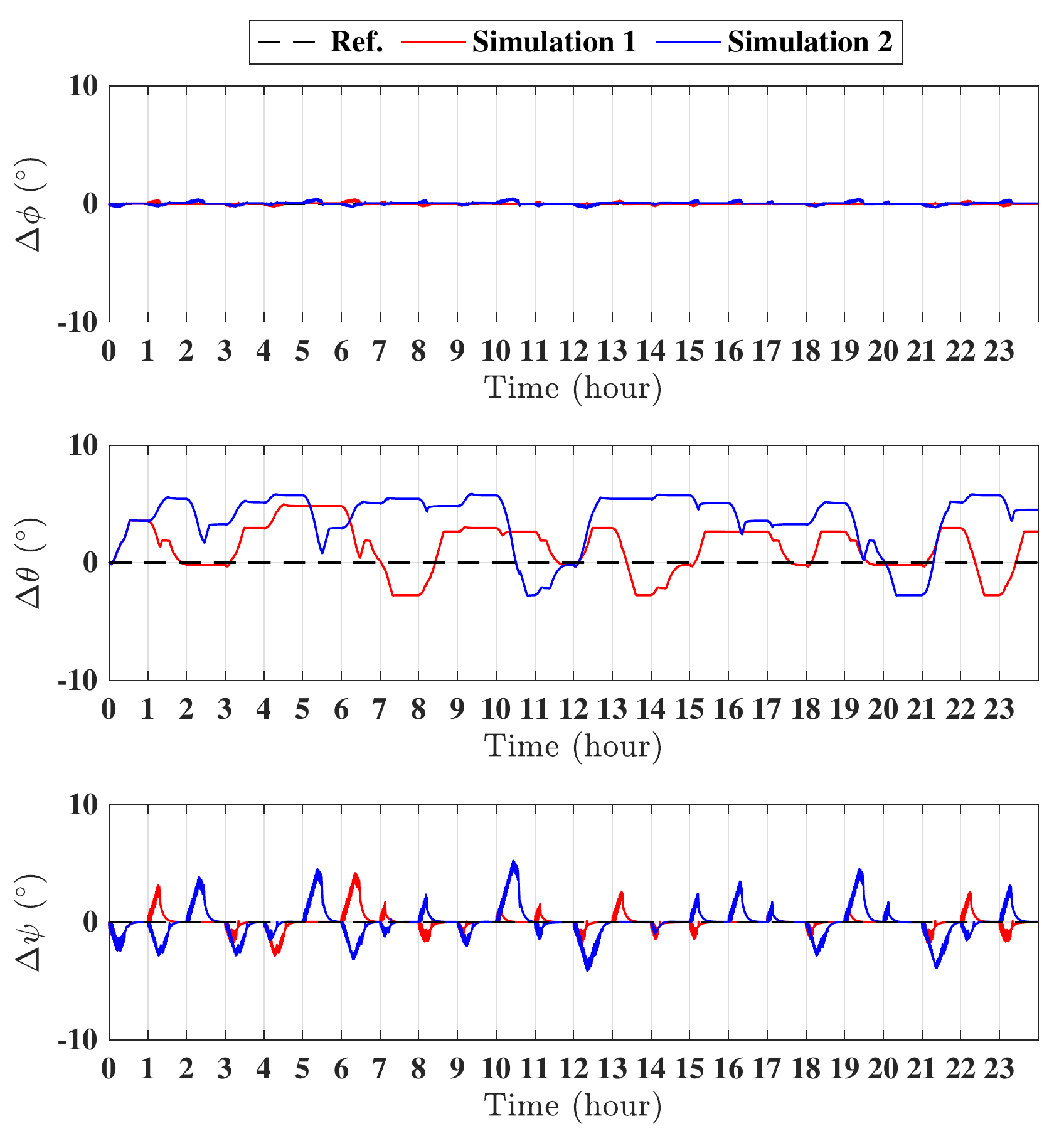}
\caption{Euler angles of the MCT system for two ocean shear profiles.}
\label{fig:euler angles}
\end{figure}

\begin{figure}[t]
\centering
\includegraphics[width=0.95\linewidth]{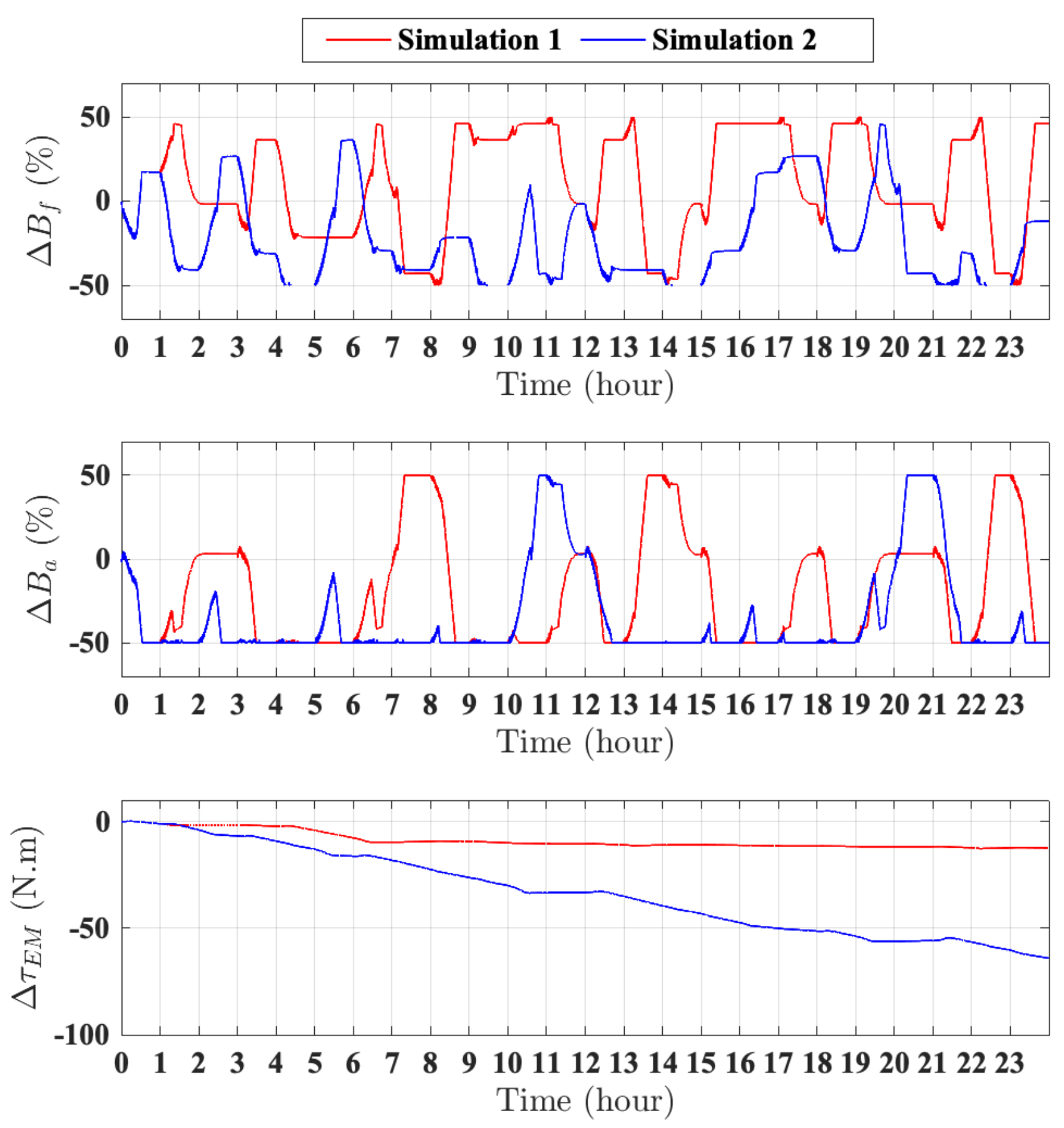}
\caption{Control inputs applied to the MCT system to track the reference trajectory for two ocean shear profiles.}
\label{fig:control_input}
\end{figure}

The goal is to maximize the net power of the MCT through modifying the operating water depth $z$ (i.e., vertical path planning), where the tracking controller tracks the optimal depth with the minimized error. After the path planning, the $z$ trajectory is sent to the tracking controller, where the error is minimized through (\ref{OF_MPC}). Fig. \ref{fig:trajectory} shows the trajectory followed by the MCT along with the optimal reference trajectory for two shear profiles, verifying that the proposed path tracking controller successfully follows the assigned optimal path by the path planner for the entire trajectory. It is shown that the path planner is able to find the optimal trajectory over low and high shear profiles, where the MCT also navigates the commanded reference paths. Results show the efficiency of the path planner in finding the optimal path to maximize the net power, where the total harvested power after 24 hours shows an over 9 \% increase compared to the baseline case (the MCT maintains at $z_{eq}=50 \,m$ without path planning). Fig. \ref{fig:euler angles} demonstrates that the Euler angles of the MCT remain within the acceptable bounds ($\pm 6^{\circ}$), and Fig. \ref{fig:control_input} illustrates the control inputs applied to the MCT to track the reference trajectory. We can see that the control inputs do not violate the constraints presented in (\ref{constraint_controller input}).

\section{CONCLUSIONS} \label{sec:conclusion}
In this paper, we presented an integrated path planning and tracking control framework for the MCT wherein the planner was designed using a DRL approach to develop a strategy in order to optimize a cost criterion (maximizing the harvested power from the MCT). The tracking control was addressed through an MPC approach considering the linear model of the system obtained from the nonlinear dynamic model. The proposed approach was applied to a buoyancy controlled MCT, and the numerical results verified that our proposed approach was able to maximize the net power through path planning compared with the baseline. Results also demonstrated that the tracking controller could effectively track the optimal path by considering the sluggish dynamics while avoiding any aggressive motion and instability.

\bibliography{root}
\bibliographystyle{ieeetr}



\end{document}